\documentclass[nohyperref]{article}

\usepackage{microtype}
\usepackage{graphicx}
\usepackage{subfigure}
\usepackage{booktabs} 

\usepackage{hyperref}

\usepackage[accepted]{icml2022}

\usepackage{amsmath}
\usepackage{amssymb}
\usepackage{mathtools}
\usepackage{amsthm}
\usepackage{multirow}

\usepackage[capitalize,noabbrev]{cleveref}

\theoremstyle{plain}

\theoremstyle{definition}

\theoremstyle{remark}

\usepackage[textsize=tiny]{todonotes}

\icmltitlerunning{Evaluating the Robustness of Real-World Super Resolution}

\begin{document}

\twocolumn[
\icmltitle{How Real is \textit{Real}: Evaluating the Robustness of Real-World Super Resolution}



\icmlsetsymbol{equal}{*}

\begin{icmlauthorlist}
\icmlauthor{Athiya Deviyani}{equal,sch}
\icmlauthor{Efe Sinan Hoplamaz}{equal,sch}
\icmlauthor{Alan Savio Paul}{equal,sch}
\end{icmlauthorlist}

\icmlaffiliation{sch}{School of Informatics, University of Edinburgh, Scotland, United Kingdom}

\icmlkeywords{Machine Learning, ICML}

\vskip 0.3in
]



\printAffiliationsAndNotice{\icmlEqualContribution} 

\begin{abstract} 
Image super-resolution (SR) is a field in computer vision that focuses on reconstructing high-resolution images from the respective low-resolution image. However, super-resolution is a well-known ill-posed problem as most methods rely on the downsampling method performed on the high-resolution image to form the low-resolution image to be known. Unfortunately, this is not something that is available in real-life super-resolution applications such as increasing the quality of a photo taken on a mobile phone. In this paper we will evaluate multiple state-of-the-art super-resolution methods and gauge their performance when presented with various types of real-life images and discuss the benefits and drawbacks of each method. We also introduce a novel dataset, WideRealSR, containing real images from a wide variety of sources. Finally, through careful experimentation and evaluation, we will present a potential solution to alleviate the generalization problem which is imminent in most state-of-the-art super-resolution models.
\end{abstract} 

\section{Introduction}
\label{sec:intro}
Super-resolution is a process of reconstructing high-resolution images from their low-resolution counterparts. However, it is well-known that super-resolution is an ill-posed problem as a large amount of current state-of-the-art methods are trained only to reconstruct images artificially downsampled by a simple and uniform degradation \cite{dong2014learning, tai2017memnet, tong2017image}. One of the most commonly used downsampling methods is bicubic downsampling. While these state-of-the-art methods produce impressive results on bicubically downsampled images \cite{ESRGAN, dai2019second}, they perform poorly on low-resolution images downsampled by an unknown, realistic image degradation method \cite{RealSR, lugmayr2019unsupervised}. Such realistic and natural degradations are what we are often presented with through cameras, smartphones, TV, etc. 

One of the reasons why natural image super-resolution is a big challenge is that there is a limited number of natural low-resolution and corresponding high-resolution image pairs. Therefore, methods such as bicubic downsampling are employed to artificially generate LR-HR pairs \cite{asurvey}. Other work has also involved using synthetically generating real low-resolution to high-resolution pairs through unsupervised learning or blind kernel estimation \cite{unsupervisedRW, zhou2019kernel}. Others have shifted their focus away from generating LR-HR pairs to simulating more complex image degradation models \cite{lugmayr2019aim}. These methods have proven to improve current super resolution models that were trained on only bicubically downsampled images to generalise better to images found in ‘the wild’. Supervised methods for “in-the-wild” real images have seen recent significant advancements compared to unsupervised methods, which are more practical in the real-world setting as it is difficult to obtain real LR-HR image pairs. We aim to perform detailed analysis of existing state-of-the-art methods to develop a better understanding of their strengths and weaknesses.

It is crucial that the trained super-resolution methods are able to generalise to real-world images, which are naturally occurring low-resolution images that have unknown and complicated downsampling kernels and noises. Examples of their applications include facial recognition in video surveillance, remote sensing, and healthcare applications such as detecting anomalies in medical images. We hope that, with this paper, we will be able to contribute to the development of SR models that are robust to the varied adverse effects (such as noises) of different equipment in these applications.

In this paper we will compare and contrast the performance of several state-of-the-art methods on real-world low-resolution images, and explore methods that could improve their generalization  performance. There are multiple metrics which indicate how well the network reconstructs the LR image to its HR version. However, calculating distortion metrics such as Peak Signal-to-Noise Ratio (PSNR) and Structural Similarity Index (SSIM) \cite{image_quality_assessment} is not possible in some cases if we are using real-world LR images for testing because no ground-truth is available. These measures are objective and do not take any subjective evaluation of human perception. We will thus use other perceptual-based metrics for qualitative evaluation on our dataset which consists of real-world images from a variety of sources.

Additionally, we note that there is a lack of datasets that contain a wide variety of naturally downsampled, real images from different types of sources. Having a wide variety of sources will help obtain more accurate measurements of generalisability of real-world  super-resolution models. Popular existing datasets such as DPED and RealSR do not achieve this. DPED contains images taken from 3 smartphone cameras, which together might have different noises compared to TV streams, CCTV footage and satellite images. RealSR contains images taken from the same camera and only helps evaluate the performance of models on the particular downsampling method of such cameras. In this paper, we aim to introduce a carefully curated dataset, WideRealSR, that can be used as a test set for thoroughly evaluating real-world performance.

In short, the goal of this paper can be summed as the following:  
\begin{enumerate}
    \item Evaluate the generalisability of various supervised and unsupervised super-resolution models,
    \item Identify the reasons why the models perform poorly or well, and 
    \item Investigate a method to devise a practical solution to potentially alleviate the generalization problem.
\end{enumerate}

\section{Dataset and task} 
As mentioned in the previous section, most of the existing datasets used in the real-world super-resolution field lack a diverse range of sensor noises. The DPED dataset \cite{dped}, for example, only contains images from 3 different smartphone cameras - iPhone 3GS, BlackBerry Passport, and Sony Xperia Z - as the Low-Resolution images. The respective High-Resolution counterparts are captured using a Canon 70D DSLR. This dataset provided support for the authors to present an end-to-end deep learning approach that bridges the gap between ordinary photos into higher-quality DSLR-like images. The authors have proposed to learn the translation function using a residual convolutional neural network that improves both color rendition and image sharpness \cite{dslr_quality}. Since its release, the DPED dataset has been the base dataset for numerous super-resolution model proposals, including the award-winning RealSR model.

A popular alternative to the DPED dataset is DIV2K \cite{div2k_dataset}, containing 800 high-resolution images and their corresponding low-resolution images that are obtained artificially through a variety of downsampling methods. Recent work has emerged \cite{realsr_and_realsrdata} in the super-resolution field which aim to identify the best image downgrading methods that best generalise to images in the real world, i.e. images with unknown sensor noises. The DIV2K dataset is commonly used in this particular field of super-resolution.

To evaluate the performance of existing super-resolution models, it is not sufficient to sample from a limited number of sensor noises and downsampling techniques. This is because, as mentioned previously, images in the ‘real-world’ inherently come with arbitrary kernels and noises. Thus, following this limitation, we decided to meticulously curate a dataset to be used for evaluating several super-resolution models. We have obtained 1-3 images for 35 different sensor noises, from sources such as (but not limited to) Google Maps, satellites, drones, microscopes, smartphones (iPhone, BlackBerry, Samsung Galaxy), WhatsApp, Facebook, tablets (iPad, Samsung Galaxy Tab), BBC broadcasts. To maintain the ‘real-world’ factor, we decided to do minimal preprocessing and only cropping the image when necessary. We call this dataset WideRealSR. We have displayed sample images from our dataset collected from 10 different sources in Figure 1.

\begin{figure*}
  \includegraphics[width=\textwidth, scale = 0.5]{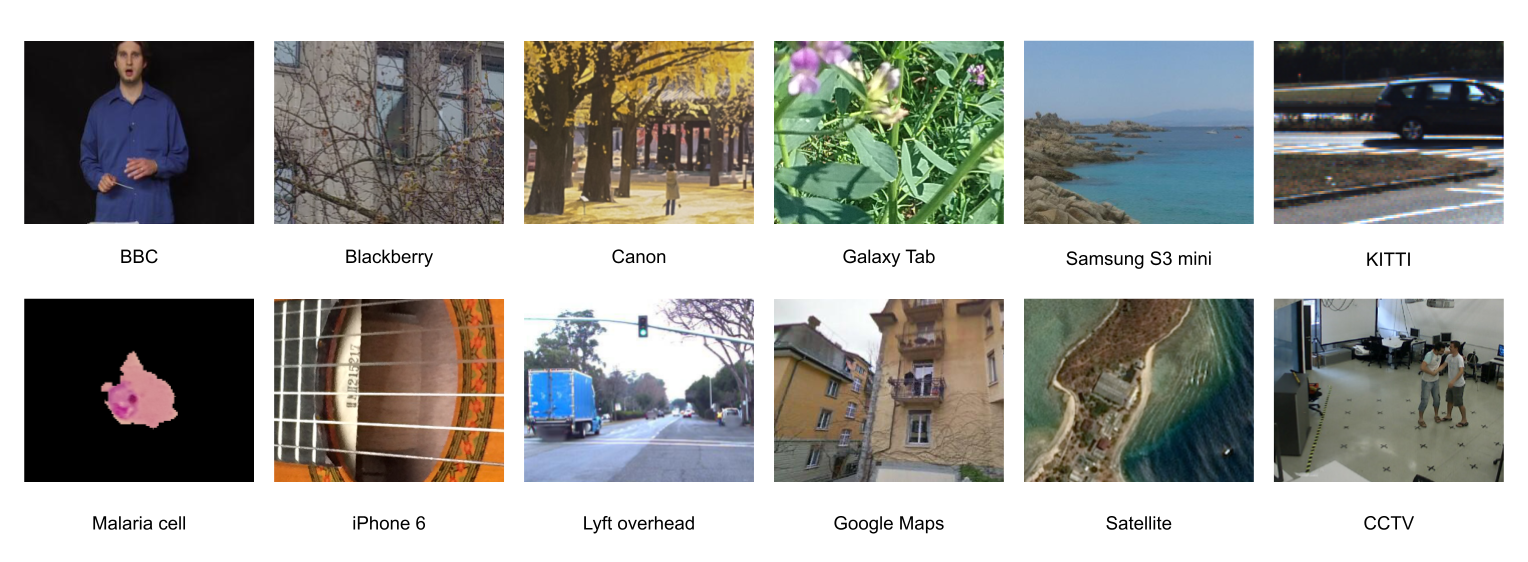}
  \caption{Sample images from the WideRealSR dataset}
\end{figure*}

The main task in this paper is to perform real-world super-resolution using state-of-the-art models on images obtained with a diversity of sensor noises. From this, we will be able to identify which models generalise best, and thus performs real-world super resolution to a satisfactory level. Due to the lack of ‘ground-truth’ high-resolution images, we will not be able to use quantitative metrics such as PSNR to evaluate the performance of the models. Instead, we decided to perform an extensive user study to obtain a human rating on the super-resolved images, which was also done for \cite{bicubic_realsr}. The details of the survey will be discussed in the proceeding sections.

\section{Methodology}
\subsection{Real-World Super Resolution Methods}

\textbf{RealSR}

The authors of the RealSR paper \cite{RealSR} propose a novel degradation framework based on kernel estimation and noise injection. The method they propose is mainly divided into two stages; the first stage is to estimate the degradation from real data and generate realistic LR images, the second is to train the SR model based on the constructed data. 
They take two sets of images as input: a real-world image set $X$, and an HR image set $Y$. They estimate the kernel (using KernelGAN \cite{kernelgan}) and noise from the $X$ set, which is then applied for downsampling the HR set to acquire an LR image set.

They then train a modified version of ESRGAN (super-resolution network, described below) with a patch discriminator, which is capable of producing HR results with better perceptual qualities. The patch discriminator helps avoid the addition of artifacts.  Compared with previously existing methods, the RealSR model produces much fewer noise and artifacts, implying that the noise estimated by noise injection is closer to the real noise. 

Their experiments on real-world images from the DPED dataset show that their model outperforms state-of-the-art methods such as EDSR \cite{edsr}, ESRGAN \cite{ESRGAN}, ZSSR \cite{zssr}, and K-ZSSR, resulting in lower noise and better visual quality. They also won the NTIRE 2020 challenge by a significant margin when scored using human perception and the Perceptual Index (PI). This is why we are keen to investigate RealSR's performance on our WideRealSR dataset.

\textbf{USRNet} 

The authors of the USRNet model proposed a deep unfolding super-resolution network. Inspired by the flexibility of model-based methods and advantages of learning-based methods, this model aims to get the ‘best of both worlds’. The main novelty aspect of the model is that it is the first paper that has successfully attempted to handle the degradation model via a single model. The core idea of this model is the deep unfolding image restoration method: it unfolds the MAP (maximum a posteriori) inference through the semi-quadratic splitting algorithm and a mixed number of iterations, consisting of alternately solving data subproblems and obtaining the prior subproblems. These two subproblems can be solved by the neural module to obtain a trainable end-to-end iterative network.

The USRNet is obtained through unsupervised learning. During training, the model takes in a high-resolution image and obtains the low-resolution image through a specially designed degradation model, which takes into account three key user-defined inputs: scale factor, the blur kernel, and the noise level. After a series of thorough experimentation, the authors were able to show that the USRNet prevailed over its competitors (such as RCAN \cite{rcan}, ZSSR \cite{zssr}). Thus, they concluded that USRNet has a high potential when used in a real-world setting. 

While the experimental results look promising, it is interesting to evaluate the network’s potential on our WideRealSR dataset containing images with an inherently unknown downsampling method, as the model relied heavily on computationally generated low-resolution images and their high-resolution counterpart during training.

\textbf{DPSR}

The Deep Plug-and-Play Super-Resolution model \cite{dpsr}, also known as DPSR. The novelty of this model lies in the image restoration method: it treats each restored model as a flexible plug-and-play (hence, the name) toolbox. The main objective of the paper is to build a new degradation model that will allow existing blind deblurring methods to estimate the blur kernel from an out-of-distribution low-resolution image. This is done in hopes to improve the robustness of existing super-resolution models in real-world applications.

Similar to USRNet \cite{usrnet}, the authors of DPSR introduce an energy function to optimize the new degradation model. The plug-and-play models are obtained through a semi-quadratic splitting technique. From this, they were able to plug any super-resolver prior rather than the denoiser prior as a modular part. The model was then evaluated through comparison between artificially generated images and real images.

Through a wide range of experiments, the authors have successfully shown that their model allowed them to insert the super-parser prior in the plug-and-play framework in the place of the denoising prior. This signifies that they are able to use the advantages of the existing neural-network-based single-image super-resolution moles to pre-design and train the super decomposer. The experiments also show that the proposed degradation model can potentially solve the super-resolution problem of arbitrary fuzzy kernels and can be used in a real-world setting to produce satisfactory results. We were curious as to what extent the new degradation model affects the generalisability of the model, thus we decided to include this model in our experiments.

\subsection{Non-Real-World Super Resolution Methods}

\textbf{ESRGAN} 

This super-resolution model \cite{ESRGAN} is an enhanced version of a previous implementation - SRGAN \cite{srgan}. The general architecture design of the network is retained but few new concepts are added and changed which ultimately lead to increase in efficiency of the network. To do this, the authors studied three key components of SRGAN: network architecture, adversarial loss, and perceptual loss, and improved each of them to derive an Enhanced SRGAN (ESRGAN).

The network structure of the generator is improved by introducing the Residual-in-Residual Dense Block which replaces the original basic blocks in SRGAN and removing all Batch Normalization layers, which contributed to the reduction of computation complexity and increase in performance. The authors have managed to improve the adversarial loss by employing a relativistic discriminator that probabilistically judges ‘whether one image is more realistic than the other’ instead of ‘whether one image is real or fake’. Finally, the authors have introduced perceptual loss to optimize super-resolution models in feature space instead of pixel space, which leads to brightness consistency and texture recovery. 

The final model that was proposed in the paper was trained as an unsupervised learning task. It creates the LR-HR pairs by bicubic downsampling the training HR images (obtained from DIV2K). It is trained to generate high-resolution images from low-resolution test images with an up-scaling factor of 4. The model was then evaluated with the aforementioned losses. Although the ESRGAN is not intended for real-world applications, it is a common backbone for models like RealSR, and models that aim to solve the real-world super resolution problems such as face images from surveillance cameras \cite{realsr_surveillance}. 

\textbf{EnhanceNet}

A common problem in objective metric-based Single Image Super-Resolution CNN models is the generation of over-smoothed images, regardless of high PSNR values. EnhanceNet \cite{enhancenet} aims to address this problem by focusing primarily on generating realistic textures along with higher perceptual quality rather than just optimizing PSNR values.

The EnhanceNet is built in the likes of a generic CNN architecture. The feed-forward fully convolutional network contains 10 residual blocks to help faster model convergence. Upsampling is done through nearest neighbor upsampling followed by a convolution layer instead of the convolution transpose layers. To ensure training stability and avoid any unwanted colour shifts, a bicubic interpolation of the low-resolution input image is added to the reconstructed output. The authors evaluated the performance of the model when implemented with a variety of combinations of loss functions, including the pixel-wise MSE-loss in the image domain, the perceptual loss in the feature space, the texture matching loss, and the adversarial loss.

The authors have presented through thorough experimentation that EnhanceNet optimized with a combination of perceptual loss, adversarial loss and texture loss generates an image that is very close to the original high-resolution image. Unlike the models presented previously, EnhanceNet aims to solve only a specific problem that is common in real-world image super-resolution model design attempts. Investigating the performance of this model on our WideRealSR dataset will help us identify whether or not alleviating the over-smoothing problem is an important factor to generate more realistic high-resolution images from low-resolution input.

\subsection{Evaluation method}
As mentioned briefly in the previous section, we evaluate the model using its pretrained weights on our dataset, WideRealSR, which contains images with a diverse range of sensor noises. We have kept the image preprocessing to a minimum as we would like to keep the images as close to actual real-world images as possible.

However, due to the various sensor noises present and the unavailability of the complementary high-resolution images in the WideRealSR dataset, we were not able to quantitatively evaluate the performance of the models. Existing quantitative super-resolution metrics such as PSNR and SSIM calculate the distortion between a model-generated high-resolution image to its respective gold-standard counterpart. In addition, recent studies have proven that the PSNR and SIM metrics do not always correlate well with super-resolution performance, signifying a discrepancy between the objective evaluation and subjective human perception \cite{quality_assessment}.

As a workaround, we turned to \cite{bicubic_realsr} paper and designed a similar qualitative evaluation pipeline which relies on the human perception ability. We performed a user study in the form of a survey where a respondent would be presented with a series of super-resolved images from the WideRealSR. These images are generated from the five super-resolution models discussed previously. Out of the five unlabeled images, the respondent is required to identify one image that they think has the highest resolution while still maintaining a realistic ‘look’, i.e. the images do not contain oversaturated colors, unrealistic contrast and brightness, etc. From this user study, we were able to identify which image super-resolution models are most resistant to unknown kernels, and thus proving its robustness for real-world applications.

\section{Experiments}
\label{sec:expts}

The experiments in this section were run on an NVIDIA Geforce GTX 1060 GPU in a virtual environment under Scientific Linux version 7.8. We have utilised the official PyTorch implementations of the models discussed in the previous section.

In this section, we will perform experiments to prove or disprove the ‘robustness’ claims made by the proponents of the state-of-the-art super-resolution models, thus evaluating the feasibility of implementing them in a real-world setting. 

The first experiment we conducted consisted of several straightforward sub-experiments described as the following:

\begin{algorithm}[ht]
\begin{algorithmic}
   \STATE {\bfseries Input:} WideRealSR Real-world LR Images $lr$, set of models $m$
   \FOR{each $m_i$ in $m$}
   \STATE Load pre-trained weights $w_i$ of $m_i$, where $w$ is obtained with the optimal model hyperparameters
    \FOR{each $lr_j$ in $lr$}
    \STATE Generate high-resolution image $hr_j$ from $lr_j$ using $m_i$
    \STATE Output $hr_j$
    \ENDFOR
   \ENDFOR
\end{algorithmic}
  \caption{Experiments using WideRealSR}
  \label{alg:example}
\end{algorithm}

After performing the experiment above, we were able to visually observe the performance of each model on out-of-distribution images (images with arbitrary kernels) by physically comparing the generated high-resolution images. 

We picked one image from each domain to avoid a burdensome evaluation process. Then, we collated their corresponding generated high-resolution images from the various models to be used in our user study, to be presented to a pool of 30 respondents. We have shown a sample of the images below:

Table 1 shows the number of times where a user identified an image generated by a particular model as the ‘best’ super-resolved image.

\begin{table}[tbh]
\centering
\begin{small}
\begin{tabular}{|l|c|c|c|}
\hline
\multicolumn{1}{|c|}{\multirow{2}{*}{Model}} & \multicolumn{3}{c|}{Counts (per domain)}       \\ \cline{2-4} 
\multicolumn{1}{|c|}{}                       & Smartphone  & Google StreetView & Other        \\ \hline
RealSR                                       & \textbf{21} & 7                 & \textbf{537} \\ \hline
USRNet                                       & 5           & 9                 & 120          \\ \hline
ESRGAN                                       & 0           & 3                 & 12           \\ \hline
EnhanceNet                                   & 0           & 1                 & 6            \\ \hline
DPSR                                         & 4           & \textbf{10}       & 315          \\ \hline
\end{tabular}
\end{small}
\caption{User Study Results: Number of votes obtained by each model}
\label{tab:my-table}
\end{table}

\begin{figure}[h]
\centering
\includegraphics[width=\columnwidth]{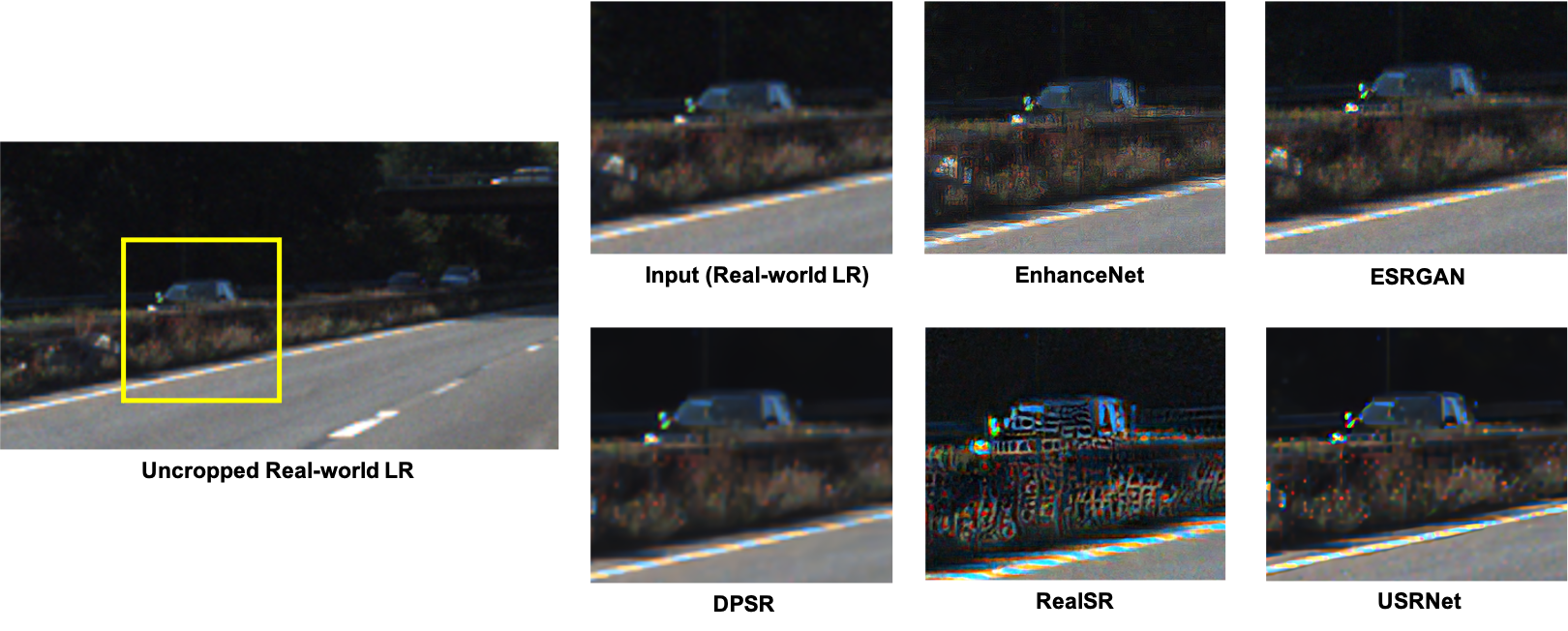}
\caption{Comparison of the performances of RealSR, EnhanceNet, ESRGAN, DPSR, and USRNet on a KITTI driving image from our WideRealSR dataset.}
\label{kitti}
\end{figure}
\begin{figure}[h]
\centering
\includegraphics[width=\columnwidth]{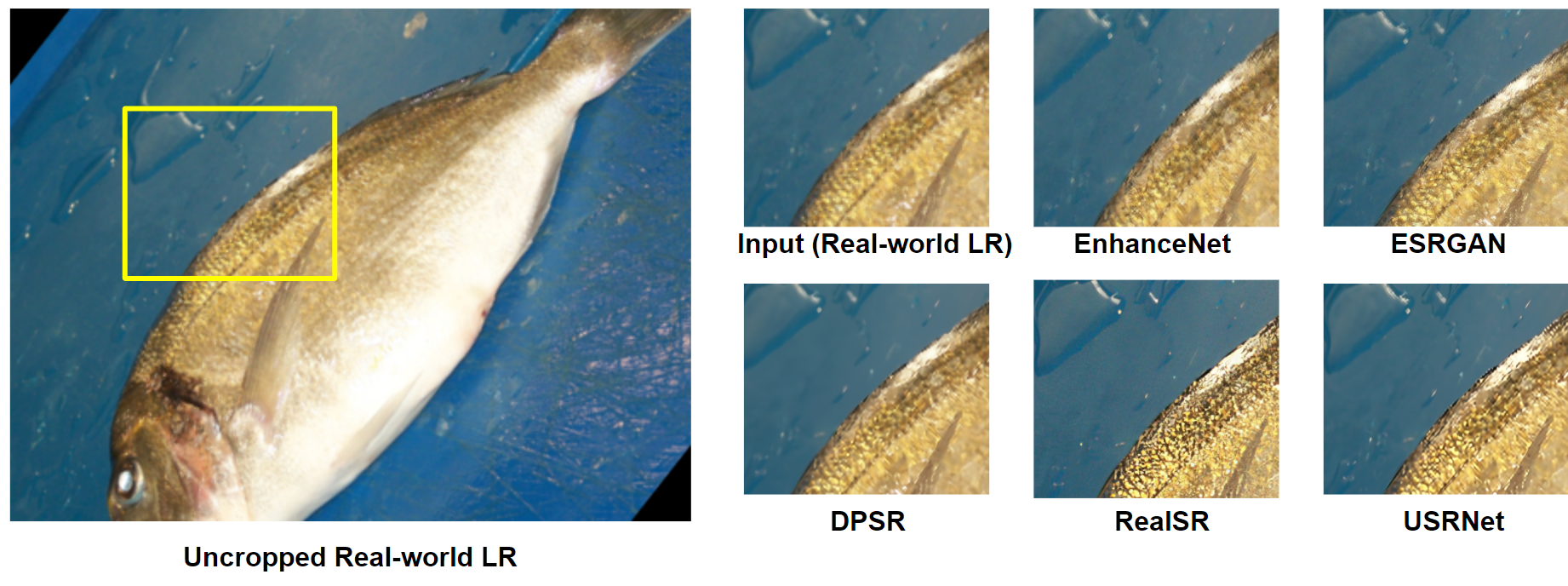}
\caption{Comparison of the performances of RealSR, EnhanceNet, ESRGAN, DPSR, and USRNet on a iPhone11 fish image from our WideRealSR dataset.}
\end{figure}

From the detailed user study, we note the inconsistency in the notably award-winning high-performing RealSR model: it performs significantly well on images taken on many smartphones, however does significantly  poorly on images captured from other sources such as Google StreetView and KITTI driving dataset. Figure 3 shows one instance where RealSR outperforms other models when evaluated on the downsampling kernel it was trained on, while Figure \ref{kitti} illustrates RealSR adding artifacts and ruining the image when evaluated on an out-of-distribution downsampling kernel. It is important to note that the RealSR model was trained on an iPhone 3GS, a BlackBerry Passport, and a Sony Xperia Z, nevertheless it performs well on sample images captured through other smartphones such as Samsung S3 mini, OnePlus 6, and Microsoft Lumia. From the results of this evaluation, we were able to formulate a hypothesis concerning the correlation between the input data sensor noise diversity to model performance: Incorporating images from a broad range of sensor noises into the training phase will improve the generalisability of the model.

However, simply generating LR-HR pairs using all sensor noises for building the training dataset would not work because the resulting model would be “confused” and super-resolve the image poorly. This is why the training dataset must be LR-HR pairs containing a single downsampling kernel. However, our goal is to make the model capable of super-resolving images with any unknown downsampling kernel. Thus, in order to do this, we propose a novel approach to training the RealSR.

In our new approach, we first group together kernels that are similar to each other using k-means clustering after finding the optimal number of clusters. Each cluster group will represent one type of kernel. For each cluster, we use the cluster centre as the downsampling kernel to create a training set (LR-HR pair) and train the RealSR model. Thus we obtain 1 model per cluster. 

For every test image, we estimate its kernel using KernelGAN \cite{kernelgan} and find which cluster it belongs to using distance from the cluster centres. We use its cluster’s RealSR model to super-resolve the LR real-world test image. Due to limited time budget, we first (1) verify that training the model on a domain apart from DPED’s sensor domain will lead to performance improvements on that domain, when compared to the performance of the model trained on the DPED domain. Then, (2) we apply K-Means clustering on the kernel groups to understand the similarities between kernels and prepare to train multiple models, which will be the last step in building a model that is more generalizable than RealSR.

To verify hypothesis (1), we first pick a poor super-resolved output image generated by the pre-trained DPED model, to compare it against our new model’s performance. 

\begin{figure}[h]
\centering
\includegraphics[width=\columnwidth]{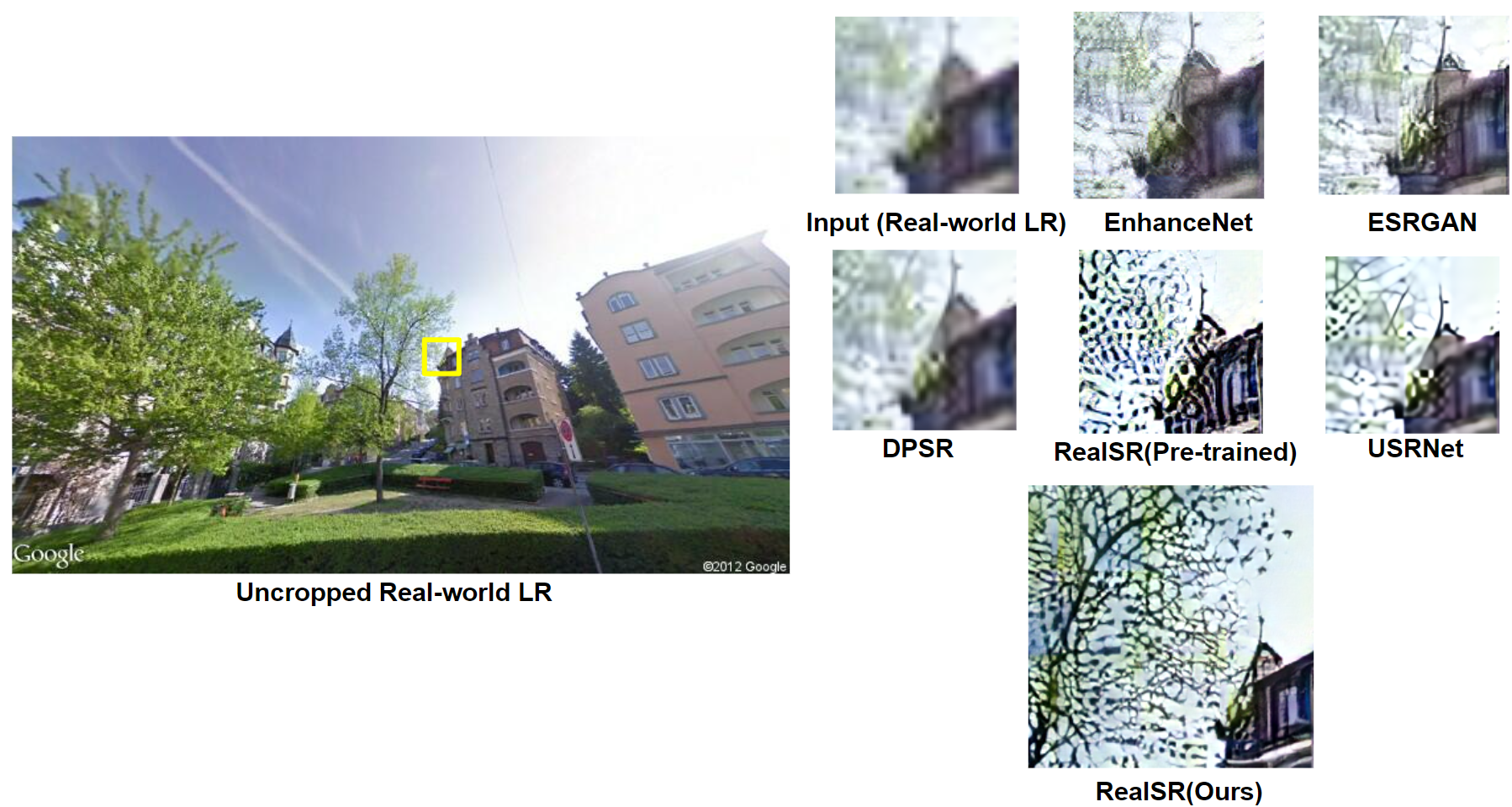}
\caption{Comparison of the performances of pre-trained RealSR, EnhanceNet, ESRGAN, DPSR, USRNet and RealSR trained on Google Street View images on a Google Street View image from our WideRealSR dataset.}
\label{selftrainresults}
\end{figure}

To train the new model from scratch, we estimate the downsampling kernel in Google StreetView images and create a train and validation dataset containing LR-HR pairs by downsampling HR images (from DIV2K \cite{div2k_dataset}) using the estimated kernel. We then train the RealSR (ESRGAN-based) model using this dataset. Details of this model are mentioned in the RealSR section above. The comparison of the performance of this model versus the pretrained models on the selected image is illustrated in Figure \ref{selftrainresults}.

The newly trained model performs significantly better than the model trained only on smartphone images. Particularly, we note that the new model has more realistic textures and does not include undesirable artifacts. 

\begin{figure}[h]
\centering
\includegraphics[width=\columnwidth]{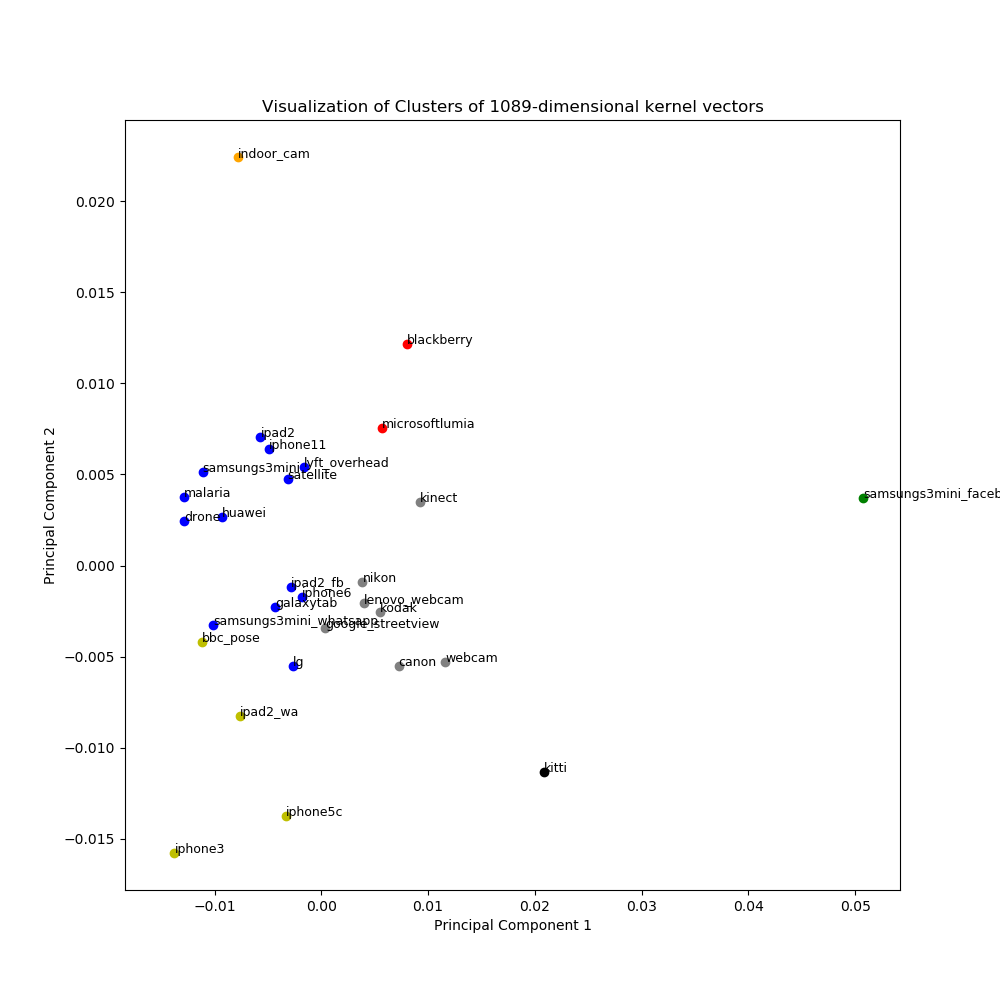}
\caption{Visualisation of dimension-reduced kernel clusters}
\label{pca}
\end{figure}

(2) With the basic hypothesis proved, we further explore the clusters of kernels that we would train models on. We flatten the kernel matrices to get (1, 1089)-shaped vectors, which we use as inputs to SkLearn’s k-means clustering algorithm with 7 clusters and illustrate the clusters in Figure \ref{pca} with further details in Table 2. For visualizing the clusters, we used PCA to convert the 1089 dimensional feature space to a 2 dimensional vector.

Thus, we successfully verified that we can make the RealSR model more generalizable by training on a dataset that has a downsampling kernel similar to the test input. We also proposed using the K-Means clustering technique to scale this improvement to arbitrary test input kernels. We leave the verification of this final proposal as potential future work.

\begin{table}[tbh]
\centering
\begin{small}
\begin{tabular}{|p{0.1\linewidth}|p{0.35\linewidth}|p{0.40\linewidth}|}
\hline
Cluster            & Source/Equipment     & Name/Names                                       \\ \hline
\multirow{9}{*}{0} & Mobile Phone         & iPhone 6, iPhone 11, LG, Huawei, Samsung S3 Mini \\ \cline{2-3} 
                   & Satellite camera     & Satellite                                        \\ \cline{2-3} 
                   & Camera               & Kodak                                            \\ \cline{2-3} 
                   & Microscope           & Malaria                                          \\ \cline{2-3} 
                   & Car camera           & LYFT-overhead                                    \\ \cline{2-3} 
                   & Tablet camera        & Galaxy Tab, iPad2                                \\ \cline{2-3} 
                   & Facebook compression & ipad2\_fb                                        \\ \cline{2-3} 
                   & Whatsapp compression & samsungs3mini\_whatsapp                          \\ \cline{2-3} 
                   & Drone camera         & Drone                                            \\ \hline
\multirow{4}{*}{1} & Camera               & Nikon, Kodak, Canon, Kinect                      \\ \cline{2-3} 
                   & Laptop webcam        & Lenovo\_webcam                                   \\ \cline{2-3} 
                   & Google Maps          & Google Street View                               \\ \cline{2-3} 
                   & Webcam               & webcam                                           \\ \hline
\multirow{3}{*}{2} & TV Broadcast         & BBC Pose                                         \\ \cline{2-3} 
                   & Whatsapp compression & iPad 2\_wa                                       \\ \cline{2-3} 
                   & Mobile Phone         & iPhone3, iPhone 5c                               \\ \hline
3                  & Mobile Phone         & Blackberry, Microsoft Lumia                      \\ \hline
4                  & CCTV                 & indoor\_cam                                      \\ \hline
5                  & Facebook compression & Samsungs3mini\_facebook                          \\ \hline
6                  & FLIR Camera          & KITTI                                            \\ \hline
\end{tabular}
\end{small}
\caption{Cluster labels and noise sources of kernels illustrated in Figure \ref{pca}}
\label{tab:my-table}
\end{table}

\section{Related work}
\label{sec:relwork}
There has not been an extensive amount of research done on comparing the overall generalisability of available state-of-the-art super-resolution models. Most of the existing work that evaluates the performance of these models are very domain specific, such as in biomedical optics for retinal imagery \cite{sr_retinal}. Similar papers would also investigate the corresponding hardware to take into consideration during the evaluation process.

On the contrary, there have been plenty of attempts to create datasets to train and evaluate super-resolution models, as we did with WideRealSR, except none of them contain a wide range of downsampling methods. A rather popular dataset is the NTIRE dataset \cite{ntire2019}. The authors introduced a novel large dataset for video deblurring, video super-resolution and evaluation of state-of-the-art models. Originating from the 2019 NTIRE challenges, the video deblurring and video super-resolution challenges are the first challenges of its kind receiving numerous participants and proposed solutions.  Another attempt of collecting datasets for conducting image super-resolution experiments is the Wieldfield2SIM \cite{w2s} dataset. Unlike NTIRE, this dataset is more domain-specific, as it consists of 100K+ real fluorescence microscopy images, containing noisy LR images with different noise levels, noise-free LR images with corresponding high-quality HR images. This dataset was used to benchmark the combinations of 6 denoising methods and 6 super-resolution methods. 

Most of the methods proposed to solve the generalization models commonly involve improving and tweaking existing models. This is done by selectively choosing different aspects of the super-resolution method to amend, such as the downsampling method, image reconstruction, and more. Some of these improvements were implemented by the models that we have discussed in the preceding sections in this report.

\section{Conclusions}
\label{sec:concl}
In this paper, we found that the models are not as generalizable as they claim, by testing them on a wide variety of sensor noises in the form of a dataset that we propose, called WideRealSR. We recognise the reason for the generalization problem: inability to deal with random unknown noise because of the lack of diversity in sensor noises available for training.

We then found a simple and potentially highly effective solution to increase the generalisability of the RealSR model. While we propose a method to improve an existing model, we find it important to point out a crucial flaw in this high-performing model: the objective loss function. We discuss that human evaluation is a better metric, currently, to evaluate the model performance. However, these models have been trained on flawed objective loss functions such as PSNR and SSIM. We encourage researchers to focus on developing a reliable metric that highly correlates with perceptual quality of the HR image. 

We would like to encourage researchers to completely verify our final proposal that could potentially outperform state-of-the-art real-world super-resolution methods.

\newpage

\bibliography{example_paper}
\bibliographystyle{icml2022}

\end{document}